\documentclass[10pt,twocolumn,letterpaper]{article}

\usepackage{cvpr}
\usepackage{times}
\usepackage{epsfig}
\usepackage{graphicx}
\usepackage{amsmath}
\usepackage{amssymb}
\usepackage{soul}
\usepackage{booktabs}

\usepackage{amsmath,amsfonts,bm, bbm}

\def\eqref#1{equation~\ref{#1}}

\def\1{\bm{1}}

\def\vf{{\bm{f}}}

\def\vs{{\bm{s}}}

\def\mF{{\bm{F}}}

\def\mS{{\bm{S}}}

\def\mV{{\bm{V}}}

\DeclareMathAlphabet{\mathsfit}{\encodingdefault}{\sfdefault}{m}{sl}
\SetMathAlphabet{\mathsfit}{bold}{\encodingdefault}{\sfdefault}{bx}{n}

\def\gD{{\mathcal{D}}}

\def\gY{{\mathcal{Y}}}

\usepackage{todonotes}
\usepackage{multirow}
\usepackage{gensymb}
\usepackage{pifont}
\usepackage{xcolor}
\usepackage[inline]{enumitem}
\usepackage{tabularx}
\usepackage{subcaption}
\usepackage{url}
\usepackage{array}

\usepackage[capitalize]{cleveref}
\crefname{section}{Sec.}{Secs.}
\Crefname{section}{Section}{Sections}
\Crefname{table}{Table}{Tables}
\crefname{table}{Tab.}{Tabs.}

\newcommand{\xmark}{\ding{55}}%
\newcommand{\cmark}{\ding{51}}%

\usepackage{color, colortbl}
\definecolor{LightCyan}{rgb}{0.88,1,1}
\definecolor{GAR}{rgb}{1,0.95,0.8}
\definecolor{GADER}{rgb}{0.87,0.92,0.97}

\definecolor{white}{rgb}{1,1,1}

\begin{document}

\title{Distillation-guided Representation Learning for Unconstrained Gait Recognition }

\author{\parbox{16cm}{\centering
{Yuxiang Guo\textsuperscript{1},
Siyuan Huang\textsuperscript{1},
Ram Prabhakar\textsuperscript{1},\\
Chun Pong Lau\textsuperscript{2}, 
Rama Chellappa\textsuperscript{1}, Cheng Peng\textsuperscript{1}}\\
{\textsuperscript{1}Johns Hopkins University, 
\textsuperscript{2}City University of Hong Kong}\\
{\tt\small \{yguo87, shuan124,   rprabha3, rchella4, cpeng26\}@jhu.edu, cplau27@cityu.edu.hk}
}
}

\maketitle
\thispagestyle{empty}

\begin{abstract}
Gait recognition holds the promise of robustly identifying subjects based on walking patterns instead of appearance information. While previous approaches have performed well for curated indoor data, they tend to underperform in unconstrained situations, e.g. in outdoor, long distance scenes, etc. We propose a framework, termed \textbf{GA}it \textbf{DE}tection and \textbf{R}ecognition (\textbf{GADER}), for human authentication in challenging outdoor scenarios. Specifically, GADER leverages a Double Helical Signature to detect segments that contain human movement and builds discriminative features through a novel gait recognition method, where only frames containing gait information are used. To further enhance robustness, GADER encodes viewpoint information in its architecture, and distills representation from an auxiliary RGB recognition model, which enables GADER to learn from silhouette and RGB data at training time.  At test time, GADER only infers from the silhouette modality. We evaluate our method on multiple State-of-The-Arts(SoTA) gait baselines and demonstrate consistent improvements on indoor and outdoor datasets, especially with a significant 25.2\% improvement on unconstrained, remote gait data.
\end{abstract}

\vspace{-0.5cm}
\section{Introduction}
\label{sec:Introduction}

Unconstrained biometric identification, in outdoor and far-away situations, has been a longstanding challenge~\cite{zhu2021gait,zheng2022gait,sepas2022deep,shen2022comprehensive}. RGB-based face and body recognition systems focus on learning \emph{spatially} discriminative features; however, real-world effects like challenging view angles, low face resolution, changing appearances (e.g., clothes and glasses), and long distance turbulence can significantly distort biometric information. Consequently, RGB-based recognition systems tend to perform inconsistently in remote unconstrained scenarios~\cite{li2018resound,li2019repair,weinzaepfel2021mimetics}. 

Gait analysis provides an alternative modality for human recognition by focusing on learning discriminative features in the \emph{temporal} domain. As such, it can be more robust to challenging, unconstrained situations, especially at range, and has been applied in many applications such as human authentication~\cite{benedek2016lidar}, health~\cite{del2019gait} and crime analysis~\cite{hadid2012can}, etc.

The field of gait recognition was initially developed~\cite{sarkar2005humanid,han2005individual,chen2009frame,liao2017pose,an2018improving,sepas2022deep} using traditional methods, such as template matching~\cite{bobick2001recognition,liu2007gait,zhang2010active} and model-based methods~\cite{benabdelkader2002stride,yoo2008automated,boulgouris2007human,li2008gait}, but limited by variations in scale and viewing angle and sensitivity to video quality, respectively. Deep learning (DL)-based approaches~\cite{chao2019gaitset,battistone2019tglstm,lin2021gait} have made significant advances in image and video-based recognition tasks compared to traditional methods. They are able to generate robust identity embeddings by directly processing the complex temporal information present in gait sequences. This enables effective recognition under the variabilities mentioned above, making the DL-based methods widely preferred.

\begin{figure}
    \centering
    \includegraphics[width=.4\textwidth, height=.35\textwidth]{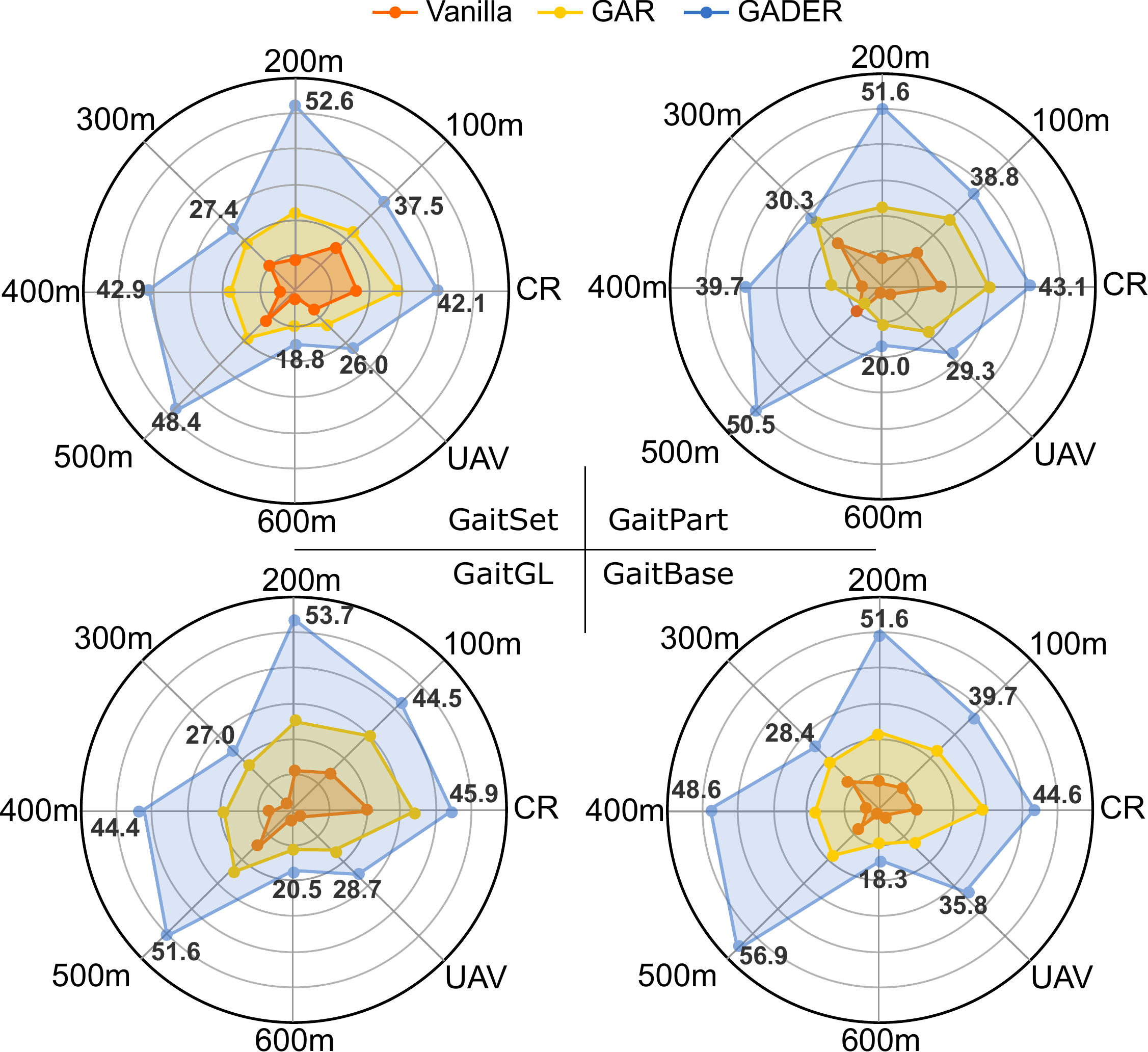}
    \caption{For each baseline, GADER raises the Rank-1 accuracy on BRIAR significantly, especially over 25\% at 500m. Close Range(CR), Unmanned Aerial Vehicle(UAV). 
    }
    \label{fig:gaitdetect}
\vspace{-0.1cm}
\end{figure}

While DL-based gait recognition performs well for indoor scenes, it often fails to achieve good performance in unconstrained/outdoor scenarios. In this work, we seek to apply gait recognition to unconstrained situations with maximal automation. 
The recently collected BRIAR~\cite{briar} dataset contains standing, structured walking, and random walking sequences, which mimic the real-world challenges in gait recognition.
Existing gait recognition methods assume that the subject is always walking with periodic movement and that there are no standing sequences~\cite{yu2006framework, takemura2018multi}. 
By making such assumptions, these methods tend to learn suboptimal representations,
sometimes only achieving 24\% on close range recognition~\cite{lin2021gait}. We also realize that standing segments inherently contain little temporal information, so it is computationally intensive to apply 3D convolutions widely used in gait recognition models. Moreover, the distinctive temporal patterns of standing and gait sequences raise a problem in generating a cohesive feature space for the same identity using one model. This highlights the need for an approach that separates frames that do not contain human motion in order to make gait recognition features more robust.

In the common application scenario of gait recognition - i.e. as a component of end-to-end video recognition, there exists a plethora of additional information at training time that is rarely considered. For instance, RGB images are required for generating human silhouettes. These RGB images contain rich information that can contribute to building robust features not captured by silhouettes alone. Intuitively, the feature space learned by a body recognition model can be used to enhance gait recognition.
Another example is viewpoint information.  Many gait recognition methods employ the practice of \textit{size normalization}~\cite{chao2019gaitset}, where the original masks are cropped and resized to the same resolution regardless of the subject's distance from the camera, leading to information loss. Particularly, such a resizing ratio over frames can implicitly offer important viewpoint information, which is useful for generating effective embeddings~\cite{chai2022lagrange}. Unfortunately, this cue is lost in the resizing operation. Based on our experiments, we show that viewpoint information helps to build a robust representation, especially in unconstrained situations.

In this paper, we aim to push the performance of unconstrained gait analysis through a framework named GAit Detection and Recognition system in the wild (GADER). 
To address the problem of mixing the moving and standing segments in sequences, we introduce a novel \emph{gait detection} module (\cref{fig:gaitdetect}) that detects the walking and non-walking parts in a sequence, so that gait recognition can just exploit the frames that contain human movement. Instead of using a 3D volume to capture the movement, our gait detector uses the Double Helical Signature~\cite{ran2010applications,niyogi1994analyzing}, a 2D pattern, with a lightweight classification model to segment the walking portion of the input sequence. Thus, we do not provide the entire sequence to the model. Instead, we split the given video into multiple windows of varying lengths to get predictions, followed by Non-Maximum Suppression, to localize the movement duration. This provides a relatively pure gait sequence for gait recognition models, yielding a robust representation and making it suitable for real-world scenarios.

For the gait recognition module, namely GAR, along with the size normalized silhouettes capturing the temporal and body shape information, we further introduce a \emph{cross-modality feature distillation} step, where we guide the intermediate gait features to be more expressive by making them close to features generated from RGB frames. This enables the gait features to maintain their robustness to appearance changes, while also benefiting from the discriminative power of RGB features. As the augmented gait features can be obtained from silhouettes and do not require the RGB frames during inference, we also gain computational efficiency. Additionally, we embed the resizing ratio from the original frame as an attention signal. This \emph{ratio attention} helps to preserve the viewpoint information that is beneficial for robust identity representation.

In summary, GADER makes three contributions:

\begin{itemize}[noitemsep, nolistsep, leftmargin=*]
    \item We introduce a light-weight gait detector to automatically detect frames that contain human movements so that gait recognition and person ReID can cooperate efficiently.
    \item We propose a novel gait recognition training strategy, which leverages the color space and size information during training; specifically, knowledge distillation on RGB features is used to enhance silhouette features' capacity. 
    \item We conduct a series of evaluations, i.e. rank retrieval and verification on CASIA-B, Gait3D and BRIAR datasets, showing consistent improvement in applying SoTA gait recognition backbones.
\end{itemize}

\section{Related Work}
\label{sec:related_work}

\begin{figure*}[htb]
    \centering
    \includegraphics[width=.95\textwidth]{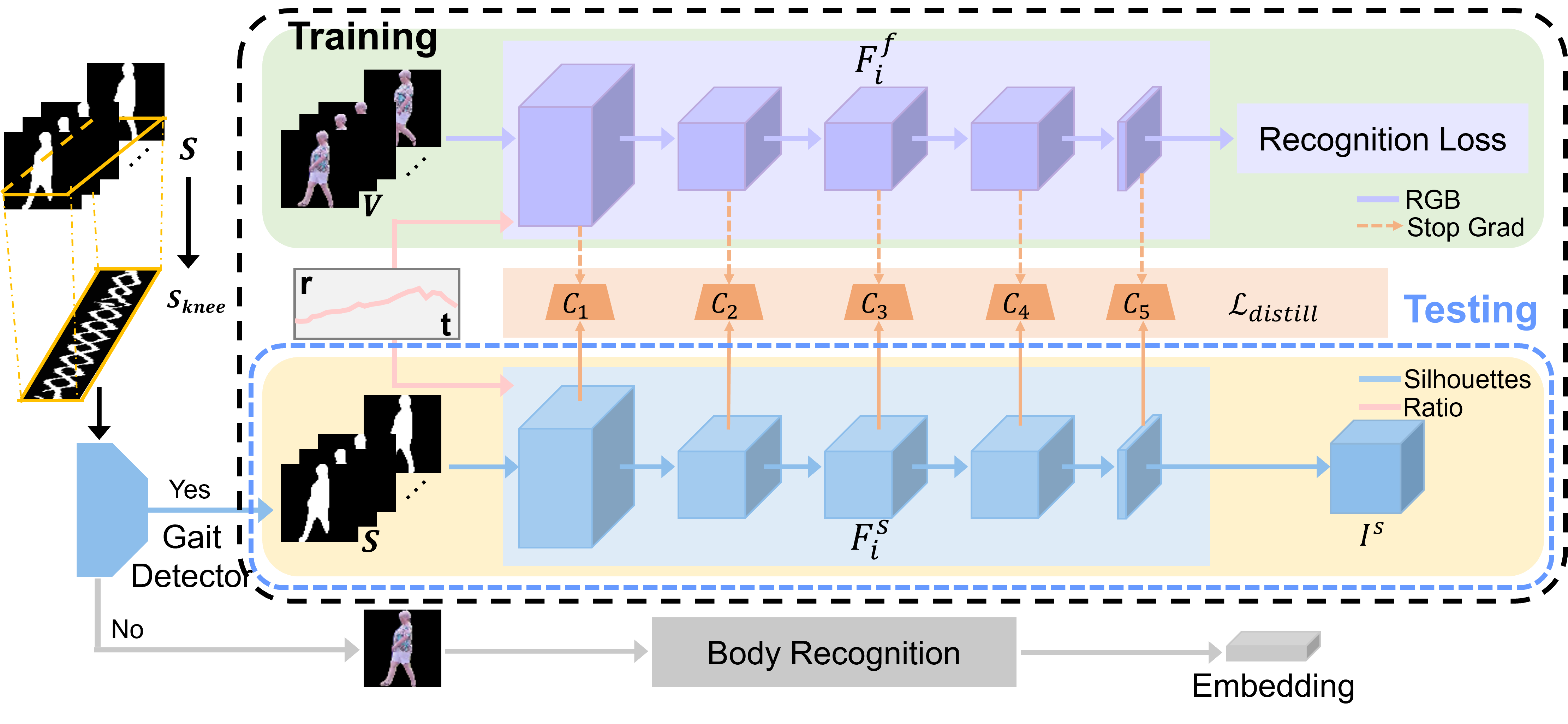}
    \caption{Overview of end-to-end pipeline. GADER consists of two parts: gait detection and GAR. The gait detector utilizes gait representation to filter out segments without gait information or incomplete body which will be processed by a body recognition algorithm, and only frames with human movement are fed to GAR. GAR leverages ratio attention and RGB feature space to extract a more robust silhouette feature for recognition.}
    \label{fig:pipeline}
\end{figure*}

\subsection{Gait Representations}
Gait, encompassing both spatial and temporal information, offers various avenues for representation, primarily classified into \textit{appearance-based} and \textit{human model-based}. Preceding the advent of deep learning, many appearance-based representations~\cite{sarkar2005humanid,chen2009frame,zhang2010active} sought to compress temporal information into a single frame. 
Han and Bhanu \cite{han2005individual} introduced the Gait Energy Image (GEI) template as an average of aligned and normalized silhouette frames. 
Other popular appearance-based gait representations include Frame Difference Energy Image \cite{chen2009frame} and Active Energy Image \cite{zhang2010active}. 
On the other hand, model-based methods represent the whole human body using well-defined models to represent gait. The methods vary by the different techniques used for modeling the human body, such as hidden Markov models~\cite{kale2004identification,liu2006improved}, stride length and walking tempo \cite{benabdelkader2002stride}, 
and Velocity Hough Transform \cite{nash1997dynamic}. 

\subsection{Gait Recognition}

Early \textit{appearance-based} methods employed global gait representation such as silhouettes \cite{chao2019gaitset}, RGB~\cite{guo2022multi, zhang2020learning, li2020end} and GEI \cite{shiraga2016geinet,wu2016comprehensive,hossain2013multimodal} as input to CNNs. The advent of deep learning technology propelled silhouettes to the forefront, primarily owing to their simplicity, privacy, and discriminative capabilities.
Dou \textit{et al.} introduced GaitMPL \cite{dou2022gaitmpl}, a progressive method that learns from simple to hard samples. 
Recently, \cite{ma2023dynamic} proposed a dynamic aggregated network (DANet) to represent contextual relationships by encoding pixel features into magnitude and phase. 
Compared to appearance-based methods, \textit{model-based} approaches~\cite{fu2023gpgait} often yield smaller input sizes extracted by lightweight models, resulting in reduced computational overhead. GaitGraph~\cite{teepe2021gaitgraph} and GaitGraph2~\cite{teepe2022towards} represent pose keypoints as a graph and extract features through Graph Convolutional Network (GCN). GaitTR~\cite{zhang2023spatial} and GaitMixer~\cite{pinyoanuntapong2023gaitmixer} employ transformer architecture to capture global temporal and spatial relationships. 
3D representations extracted using depth sensors \cite{hofmann2014tum,nunes2019benchmark,chattopadhyay2015frontal} and RGB images \cite{liao2020model,liao2017pose,cao2017realtime,kolotouros2021probabilistic,georgakis2020hierarchical} also have shown promising results.

\noindent\textbf{Recognition in the Wild}: Developing an accurate gait representation in the wild is a long-term goal and has been actively researched over the past decade. Towards that aim, Zhu \textit{et al.} curated a large scale gait dataset called GREW \cite{zhu2021gait}. It is a natural video dataset consisting of 128K sequences of 26K identities captured over 882 cameras. 
Similarly, Zheng \textit{et al.} \cite{zheng2022gait} collected Gait3D dataset that contains silhouettes, 2D/3D keypoints, and 3D meshes for 3D gait recognition. The authors~\cite{zheng2022gait} observed that state-of-the-art gait recognition methods do not yield similar superior performance for GREW and Gait3D as they do for indoor datasets like CASIA-B.

\section{Method}
\label{sec:Method}

In this work, we focus on silhouette-based gait recognition, which relies on the binary masks of the subjects in a video. Formally, we denote a video as a 4D tensor, i.e. $\mV \in \mathbb{R}^{T \times H \times W \times 3}$, where $T, H, W,$ are the frame index, height and width. For each frame $t$, the subject silhouette $\vs_t \in \{0, 1\}^{H \times W}$ is obtained from an off-the-shelf segmentation model, e.g.~\cite{wu2019detectron2}. Gait recognition takes $\mS = [\vs_t]_{t=1}^T$ as the input, 
and obtains corresponding features $\vf = F_\theta(\mS) $ from a feature extractor $F_\theta$. Triplet loss~\cite{DBLP:journals/corr/HofferA14} is used to constrain the training process of $F_\theta$ with respect to ground-truth labels $y \in \{ 1, 2, \dots, |\gY| \}$ for each video in the training sets, where $|\gY|$ is the cardinality of the label set. After $F_\theta$ is trained, gallery gait silhouettes $\mS^g$ and probe gait silhouettes $\mS^p$ are passed through $F_\theta$ to obtain the gait feature $\vf^g$ and $\vf^p$. To recognize the probe identity, a similarity metric $\gD$, such as Euclidean distance or cosine similarity, is used to get $\gD(\vf^g, \vf^p) $, where the gallery subject $g$ and the probe $p$ are decided to be the same person if they are the close enough in feature space.

\subsection{Gait Detector}
Previous gait recognition methods~\cite{chao2019gaitset,fan2020gaitpart,lin2021gait,liang2022gaitedge,chai2022lagrange} directly use the silhouettes $\mS$ as the input, implicitly under the assumption that the video sequence captures \emph{the entire body} and \emph{continuous movement}. While these assumptions are likely to be effective for curated datasets, such as CASIA-B~\cite{yu2006framework}, 
they are often ineffective for unconstrained videos and will lead to suboptimal performance~\cite{gaitquality}. To this end, we propose a gait detector to assess if the video sequence contains gait movement with the complete body, analogous to the role played by face detection in face recognition, retaining only those frames that contain human movement for subsequent processing.

\subsubsection{Double Helical Signature}
\label{subsec:gait repre}

An ideal representation of a gait detector should be able to discriminate moving subjects from stationary ones and a partial body from a full body. 
Since the legs' motion is nearly periodic and contributes significantly to gait recognition, we use the Double Helical Signature (DHS)~\cite{ran2010applications}. DHS is a classic gait representation that captures the movement of the knee to describe gait movement as a function of time. As our input $\mS$ is normalized to the same size, we can deduce the knee height $ \mathcal{H}_{\textrm{knee}}$ to be approximately a quarter of the overall frame height. By taking a slice from the silhouette sequence, i.e., $\mS_{\textrm{knee}}(x,t) = \mS (x,\mathcal{H}_{\textrm{knee}},t), \mS_{\textrm{knee}} \in \mathbb{R}^{ W \times T }$, we obtain a DHS pattern that indicates human movement.

\begin{figure}[h]
    \centering
    \begin{tabular}[b]{cc}
        \begin{subfigure}[b]{.48\linewidth}
            \includegraphics[width=.9\textwidth,height=0.22\textwidth]{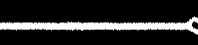}
            \caption{\footnotesize full-stand}
            \label{fs}
        \end{subfigure} &
        \begin{subfigure}[b]{.48\linewidth}
            \includegraphics[width=.9\textwidth,height=0.22\textwidth]{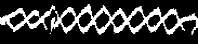}
            \caption{\footnotesize full-gait}
            \label{fg}
        \end{subfigure} \\
        \begin{subfigure}[b]{.48\linewidth}
            \includegraphics[width=.9\textwidth,height=0.22\textwidth]{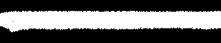}
            \caption{\footnotesize part-stand}
            \label{ps}
        \end{subfigure} &
        \begin{subfigure}[b]{.48\linewidth}
            \includegraphics[width=.9\textwidth,height=0.22\textwidth]{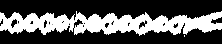}
            \caption{\footnotesize part-gait}
            \label{pg}
        \end{subfigure} \\
        
    \end{tabular} 
    \caption{Four cases of DHS(a-d) are shown using two variables: full/part - indicates whether the body is complete, and stand/gait - shows whether gait information is present.}
   \label{fig:GEIDHS}
   \vspace{-4mm}
\end{figure}

As shown in \cref{fig:GEIDHS}, the DHS pattern is discriminating. For the standing case, the DHS pattern shows a constant straight line since there is no movement at the knees. When the subject is walking, a periodic pattern is obtained, known as a type of ``Frieze pattern'' \cite{niyogi1994analyzing}. Given an incomplete body, DHS appears rather different as $\mathcal{H}_{\textrm{knee}}$ does not correspond to the knee position anymore; consequently, DHS becomes thicker. Therefore, DHS emerges as a compact and distinctive representation encapsulating temporal movement into a 2D image. 

\begin{figure}[t]
    \centering
    \includegraphics[width=.455\textwidth, height=.23\textwidth]{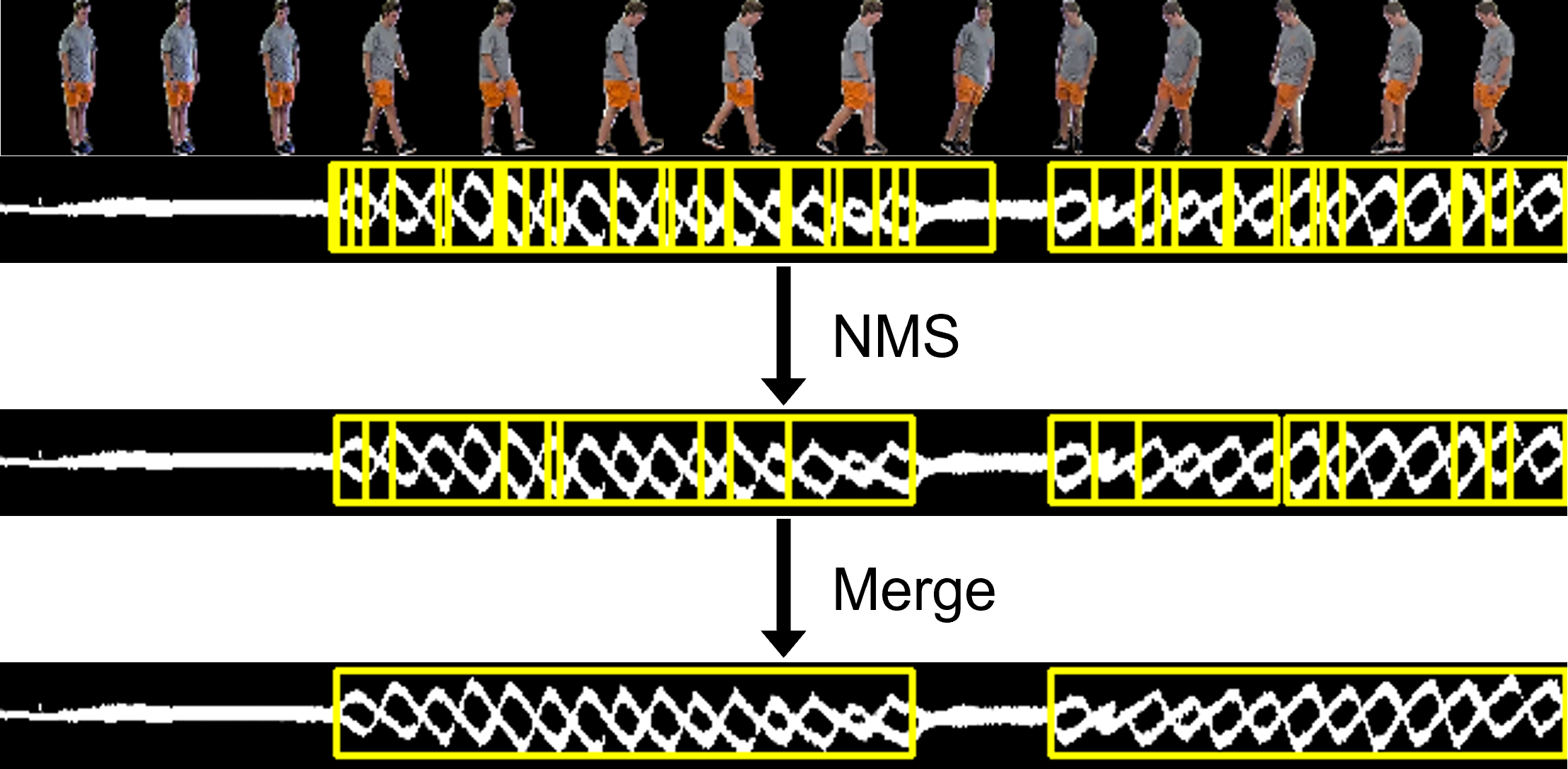}
    \caption{The gait detection process. The gait detector processes split DHS segments to obtain prediction followed by Non-Maximum Suppression (NMS) and concatenation to pinpoint frames where gait information is present in a sequence. The yellow bounding boxes indicate the frames that contain gait.}
    \label{fig:gaitdetect}
    \vspace{-4mm}
\end{figure}

\subsubsection{Light-weighted Classification}
 Capitalizing on the distinctive characteristics of DHS, we employ a simple yet effective network to extract segments from $\mS$ that contain gait information and a complete body. During the training step, we randomly select the start point and duration from the entire DHS and get $R_{\textrm{knee}}$, which is similar to the sampling strategy commonly applied in gait recognition. Each segment has the same height as DHS's. The fragment is then processed by a five-layer Convolutional Neural Network $\mathcal{M}_{\phi}$ to get the feature. To handle the varying window lengths, we employ a temporal pooling module in the form of a max-pooling layer to generate a window-length invariant embedding. Subsequently, a four-class multi-layer perceptron (MLP) is applied to obtain the prediction, using the cross entropy loss. 
\vspace{-1mm}
\begin{equation}
    \begin{split}
        P &=  \textrm{MLP}(\mathcal{P}_{Max}^{ 1\times 1\times t }(\mathcal{M}_{\phi}(R_{\textrm{knee}}))),
        \end{split}
    \label{equ:att}
\end{equation}

\noindent where $\mathcal{P}_{Max}^{ 1\times 1\times t }$ represents a temporal pooling module with size ($ 1\times 1\times t$), and $t$ is the window's width.

Especially in an unconstrained walking sequence, non-ideal fragments, including turning and standing within a short period, limit detection performance.  To precisely identify the segments that contain gait information as well as a complete boy,
we first split the entire DHS sequence into multiple windows of varying durations $R_{\textrm{knee}}^{n}$, where $n$ represents the number of windows. Each window goes through the well-trained gait classification model and gets a corresponding prediction. We only keep the complete body gait predictions and reduce the predictions by Non-Maximum Suppression (NMS). Finally, we check the reduced windows' inner distance and merge them if the distance is smaller than a predefined threshold. The ratio of the detected movement length to the entire DHS sequence serves as an indicator for determining whether the sequence should be further utilized in the gait recognition module. The process is illustrated in \cref{fig:gaitdetect}. Thus our gait detector automatically identifies the corresponding movement status of each clip.

\subsection{GAit Recognition (GAR)}
With the help of the gait detector, we can obtain relatively robust gait sequences for gait recognition. In the recognition stage, we use GAR, which incorporates prior knowledge such as the RGB feature space and the resizing information. The architecture is shown in \cref{fig:pipeline}.

\subsubsection{Ratio Attention}
\label{sec:ratio}

Viewpoint information has been shown to improve gait recognition performances~\cite{chai2022lagrange}; however, such information is difficult to obtain in unconstrained situations. In this work, we leverage resizing ratios to dynamically describe the changing views. The \emph{resizing ratio} is defined as the change in height from the original bounding box to the normalized silhouette height, which is usually $64$. Intuitively, the resizing ratios are similar if two videos are recorded from the same viewpoint, and these ratios naturally encode viewpoint information.

To effectively incorporate ratios into the gait feature extraction module, we apply it as an attention mechanism and fuse it into the network. Empirically, attention is an effective way to employ view point information as shown in \cref{sec:ablation}. The ratios are embedded using 1D convolution $F_{1DConv}$, followed by a sigmoid function $Sigmoid(*)$, i.e.
\begin{equation}
    \begin{split}
        r = Sigmoid(F_{1DConv}(\frac{\mS_{raw} }{\mS})),
    \end{split}
    \label{equ:att}
\end{equation}
where $\mS_{raw}$ and $\mS$ are the body bounding boxes in the original and normalized silhouette, and $r \in \mathbb{R}^{1\times 1\times T}$.  Notably, we utilize the ratio as a pattern, rather than a discrete representation~\cite{chai2022lagrange}, to characterize the viewing point, effectively and dynamically adapting to the viewpoint changes in unconstrained scenarios.

\subsubsection{Cross Modality Distillation}
Previous works~\cite{fan2020gaitpart,lin2021gait,liang2022gaitedge,xiao2022learning} have shown that silhouette-based recognition is robust to appearance changes such as different clothing and low image quality. On the flip side, segmenting RGB frames into silhouettes leads to loss of useful information for human identification, e.g. 
the rich content~\cite{ye2024biggait,guo2022multi}. While gait recognition systems do not have access to color space information at \emph{test time}, they can benefit by learning from the feature space learned by an RGB-based recognition system \emph{during training} to enhance the ability to separate identities. 
To this end, GAR introduces an auxiliary branch to extract features from RGB. We denote $F_{3DConv}^{f},F_{3DConv}^{s}$ as the first 3D convolution layers to the RGB and silhouette feature extraction backbones. Combining with ratio attention $r$, GAR first obtains weight features, i.e. 
\begin{equation}
    \begin{split}
        \mF^f_1 = r*F_{3DConv}^{f}(\mV)&,\mF^s_1 = r*F_{3DConv}^{s}(\mS).
    \end{split}
    \label{equ:network}
\end{equation}
Subsequently, we apply the rest of backbones to the processed features,
\begin{equation}
\begin{aligned}
    \mF^f_{i+1} = F^{f}_i(\mF^f_i)&,
    \mF^s_{i+1} = F^{s}_i(\mF^s_i),\\
    I^{f} = F^{f}_N(\mF^f_N)&,
    I^{s} = F^{s}_N(\mF^s_N),
\end{aligned}
\end{equation}
in which $I^{f}$ and $I^{s}$ are identification embedding, and $i \in \{1,2,...,N\}$ is the convolution block index.

\subsection{Loss functions}
A cross-modal distillation loss is employed within GAR, which promotes the representation power of gait features based on the learned RGB feature space. 
Since both modalities have their specific advantages, directly forcing their features to be close leads to an averaged representation that does not benefit from the specificity of each modality. An additional convolutional layer $C_i$ is introduced for the $i$-th intermediate silhouette feature $\mF_{i}^{s}$, such that the transformed features are constrained to be similar to $\mF_{i}^{f}$. The loss can be described as:

\begin{equation}
  \mathcal{L}_{distill}  = \frac{1}{N}\sum_{i}(\gD(\varnothing(\mF_{i}^{f}), C_i(\mF_{i}^{s}))),
  \label{equ:dense}
\end{equation}
 where stop gradient ($\varnothing$) operation is used such that the RGB branch is not affected by the silhouette features in this process. Note that $C_i$ is only used at training time.

Triplet loss~\cite{DBLP:journals/corr/HofferA14} $\mathcal{L}^{f}_{tri}$, $\mathcal{L}^{s}_{tri}$ is applied to maximize the distance of representations from different subjects and minimize the ones from the same identity for both modalities.

Overall, the training loss is
 \begin{equation}
  \mathcal{L}_{train}  = \lambda_{f}\mathcal{L}^{f}_{tri} + \lambda_{s}\mathcal{L}^{s}_{tri} + \lambda_{distill}\mathcal{L}_{distill},
  \label{equ:loss}
\end{equation}
where $\lambda_{f,s,distill}={0.425,0.425,0.15}$ are the loss hyperparameters used during training.

\newcolumntype{"}{@{\hskip\tabcolsep\vrule width 1pt\hskip\tabcolsep}}
\begin{table*}[t]
\setlength{\tabcolsep}{4pt}
\centering
\small
\begin{tabular}{  c |c |c |c  |c |c | c |c |c |l !{\vrule width 2pt} c |c |c |c }
\toprule
\multicolumn{10}{c!{\vrule width 2pt}}{Rank Retrieval} & \multicolumn{4}{c}{ Verification } \\
\midrule
\multicolumn{1}{c|}{Probe} & CR & 100m & 200m & 300m & 400m & 500m & 600m & UAV& Mean & $1e^{-4}$ & $1e^{-3}$ & $1e^{-2}$ & $1e^{-1}$ \\
\midrule
GaitSet~\cite{chao2019gaitset}& 21.3 & 20.8 & 13.4  &14.5 & 8.4 & 15.4 & 6.2 & 11.7 & 17.6 & 5.9 & 15.0 & 33.0 &62.2\\
\rowcolor{GAR} $\hookrightarrow$ w / GAR & 31.6 & 26.1 & 24.9& 22.7  & 21.7 & 22.3 & 14.1  & 17.1 & 26.8 \textcolor{green}{$\uparrow$ 9.2} & 9.0 & 21.2 & 41.2 & 67.5 \\

\rowcolor{GADER} $\hookrightarrow$ w / GADER & 42.1& 37.5 & 52.6  & 27.4& 42.9 & 48.4 & 18.8 & 26.0& 35.9 \textcolor{green}{$\uparrow$ 18.3} & 9.3 & 22.5 &45.7 & 75.6 \\
\midrule
GaitPart~\cite{fan2020gaitpart}  & 20.2 & 17.7 & 12.4 & 21.1  & 10.2 & 13.8 & 5.6 & 6.2 & 15.7 & 4.0 & 11.0 & 25.9 & 60.7 \\
\rowcolor{GAR} $\hookrightarrow$ w / GAR &32.3 & 30.1 & 25.8 & 29.1  & 17.8 & 11.3 & 14.7 & 21.5 & 27.3 \textcolor{green}{$\uparrow$ 11.6} & 10.3 & 21.0 & 40.9 & 74.3 \\

\rowcolor{GADER} $\hookrightarrow$ w / GADER &43.1& 38.8 & 51.6 & 30.3 & 39.7 & 50.5 & 20.0 & 29.3&  36.3 \textcolor{green}{$\uparrow$ 20.6} & 10.4 & 22.4 & 45.5 & 79.7  \\
\midrule
GaitGL~\cite{lin2021gait} & 24.6 & 18.1 & 15.5 & 7.5  & 11.0 & 18.1 & 6.2 & 6.5 & 17.3 & 4.9 & 12.9 & 33.1 & 65.9 \\
\rowcolor{GAR} $\hookrightarrow$ w / GAR & 35.8 & 32.3 & 27.8 & 21.6  & 23.0 & 27.1 & 15.2 & 19.8 & 28.3 \textcolor{green}{$\uparrow$ 11.0} & 7.1 & 16.4 & 33.6 & 60.4 \\

\rowcolor{GADER} $\hookrightarrow$ w / GADER & 45.9& 44.5 & 53.7  & 27.0& 44.4 & 51.6 & 20.5 & 28.7 &   37.8 \textcolor{green}{$\uparrow$ 20.5} & 8.2 & 19.6 & 41.1 & 71.3\\

\midrule
GaitBase~\cite{fan2023opengait} & 14.8 & 13.3 & 12.3 & 15.5 & 8.1 & 12.2 & 4.5 & 6.0 & 11.6 & 3.0 & 7.6 & 20.9 & 54.4 \\
\rowcolor{GAR} $\hookrightarrow$ w / GAR &24.0 & 22.6 & 23.7 & 23.0 & 19.1 & 22.3 & 14.1 & 17.1 & 26.8 \textcolor{green}{$\uparrow$ 15.2} & 5.5 & 13.1 & 31.6 & 65.7 \\

\rowcolor{GADER} $\hookrightarrow$ w / GADER & 44.6 & 39.7 & 51.6  & 28.4& 48.6 & 56.9 & 18.3 &  35.8 &   36.8 \textcolor{green}{$\uparrow$ 25.2} & 7.2 & 17.6 & 39.9 & 74.5 \\
\bottomrule

\end{tabular}%

\caption{Rank-1 accuracy (\%) and verification (TPR(\%)@FPR =$1e^{-4}$,  $1e^{-3}$, $1e^{-2}$, $1e^{-1}$) on BRIAR. CR is Close Range($<$100m); Unmanned Aerial Vehicle (UAV).}
\label{tab:BRIAR_rank_veri}
\end{table*}

\section{Experiments}
\label{sec:Experiments}

\subsection{Datasets and Metrics}

In this work, we focus on applying gait recognition to unconstrained scenarios with comparatively sparse data curation, exemplified by \textbf{BRIAR}~\cite{briar}. BRIAR consists of 776 and 856 subjects used as training and test sets, respectively. In the test set, there are 493 distractors and 363 target subjects. 
Compared to other unconstrained datasets, BRIAR data includes
different walking status and clothes settings, and incomplete body shapes into consideration, making it very challenging. 
Similar to previous methods for gait recognition, we also evaluate the proposed approach on \textbf{CASIA-B}~\cite{yu2006framework}, a controlled, indoor dataset with continuous motion. CASIA-B consists of 124 subjects with three walking conditions which are normal walking (NM), walking with a bag (BG) and walking in a coat or jacket (CL). 
To further demonstrate the effectiveness of the proposed approach in the unconstrained case, we also test GADER on the \textbf{Gait3D}~\cite{zheng2022gait} dataset. Gait3D consists of 25,309 sequences recorded by 39 cameras on 4,000 subjects inside a large supermarket. 

Considering CASIA-B and Gait3D are pure gait datasets, we assume that all sequences in these datasets fully contain gait, so we only evaluate \textbf{GAR}. For the BRIAR dataset, we use the proposed gait detector to extract the segments containing gait information with a complete body and feed them to the GAR model; the remaining frames are processed by a SoTA ReID method, SemReID~\cite{huang2023selfsupervised}, the resulting system is named as \textbf{GADER}.

\noindent\textbf{Evaluation Metric} For CASIA-B and BRIAR, \textit{verification} and \textit{rank retrieval} are used to evaluate recognition performance. 
As for Gait3D, the evaluation follows the open-set instance retrieval setting and calculates the average Rank-1, 5, and 10 accuracies, mean Average Precision (mAP), and mean Inverse Negative Penalty (mINP) over queries.

\begin{table*}[!htb]
\small
\centering
\begin{tabular}{ c | c |c |c |c |c |c |c |c |c |c |c |c |l }
\hline
\multicolumn{2}{c|}{Gallery NM\#1-4} & \multicolumn{12}{c}{$0\degree - 180\degree$}\\
\hline
\multicolumn{2}{c|}{Probe} & $0\degree$ & $18\degree$ & $36\degree$ & $54\degree$ & $72\degree$& $90\degree$& $108\degree$ & $126\degree$ & $144\degree$& $162\degree$& $180\degree$& Mean \\

\hline
\multirow{9}{*}{CL\#1-2} 

 & GaitSet~\cite{chao2019gaitset} & 61.4 & 75.4 & 80.7 & 77.3 & 72.1 & 70.1 & 71.5 & 73.5 & 73.5 & 68.4 & 50.0 & 70.4 \\
 \rowcolor{GAR} \cellcolor[rgb]{1,1,1} &  $\hookrightarrow$ w / GAR & 67.5 & 82.3 & 84.1 & 79.2 & 70.8 & 68.8 & 71.4 & 75.2 & 77.7 & 75.9 & 58.1 &73.7 \small \textcolor{green}{$\uparrow$ 3.3} \\

 & GaitPart~\cite{fan2020gaitpart} & 70.7 & 85.5 & 86.9 & 83.3 &  77.1 & 72.5 & 76.9 & 82.2 & 83.8 & 80.2 & 66.5 & 78.7\\ 
  \rowcolor{GAR} \cellcolor[rgb]{1,1,1} & $\hookrightarrow$ w / GAR & 78.0 & 85.7 & 89.4 & 84.4 & 77.7 & 71.6 & 77.0 & 79.7 & 84.3 & 81.3 & 68.2 & 79.7 \small \textcolor{green}{$\uparrow$ 1.0}\\

  & GaitBase~\cite{fan2023opengait}  & - &-   &- &- &- &- &- &- &- &- &- & 77.4  \\
    \rowcolor{GAR} \cellcolor[rgb]{1,1,1} & $\hookrightarrow$ w / GAR & 69.3 & 80.1 & 82.5 & 82.1& 81.2 & 78.2 & 77.8 & 80.0 & 83.3 & 80.2 & 65.3 & 78.2 \small \textcolor{green}{$\uparrow$ 0.8}\\
    
 & GaitGL~\cite{lin2021gait} & 76.6 & 90.0 & 90.3 & 87.1 & 84.5 & 79.0 & 84.1 & 87.0 & 87.3 & 84.4 & 69.5 & 83.6 \\  
  \rowcolor{GAR} \cellcolor[rgb]{1,1,1} &$\hookrightarrow$ w / GAR & 80.8 & \textbf{90.5} & \textbf{92.2} & 91.0 & 84.7 & \textbf{79.7} & \textbf{84.8} & 89.6 & 89.7 & \textbf{87.1} & \textbf{72.8} & 85.7  \small \textcolor{green}{$\uparrow$ 2.1}\\

\hline
\end{tabular}
\caption{Rank-1 accuracy (\%) on CASIA-B  excluding identical-view case for CL\#1-2.}
\label{tab:CASIA-B_Rank}
\vspace{-2mm}
\end{table*}

\subsection{Quantitative Evaluation}
\label{subsec:quan}

\noindent\textbf{Evaluation on BRIAR}~\cite{briar} To demonstrate the superior performance of the proposed methods in the wild, we evaluate on BRIAR data with GaitSet~\cite{chao2019gaitset}, GaitPart~\cite{fan2020gaitpart},  GaitGL~\cite{lin2021gait} and GaitBase~\cite{fan2023opengait} serving as backbones. 
The Rank-1 accuracies are shown in ~\Cref{tab:BRIAR_rank_veri}. 
With the help of distilled knowledge from RGB data, our gait recognition models gain improvements compared to their vanilla version, by 9.2\%, 11.6\%, 11.0\% and 15.2\%, respectively. GAR leads to better performance than previous works, indicating that the referred discriminative feature space from distillation contributes to a better gait feature. When the gait detector as well as SemReID are integrated, GaitGL with GADER doubles the vanilla's accuracy from 17.3\% to 37.8\%,  showing that it is beneficial to apply the gait detector to eliminate segments without human movements.

When it comes to \textit{verification}, the TAR (\%) results are shown 
in ~\Cref{tab:BRIAR_rank_veri}. For the case of GaitBase, compared to the original, our method improves by 10.7\%, achieving 31.6\% when FAR=$1e^{-2}$.
When the gait detector is included, we observe a big improvement. The recognition system's verification results increase from 20.9\% to 39.9\% FAR@$1e^{-2}$indicating that gait recognition will perform well if the segments that contain gait information with a complete body are used as input.

\noindent\textbf{Evaluation on CASIA-B}~\cite{yu2006framework} To show the effectiveness of the proposed method in indoor scenarios, we present the 
performance variations across four backbones by incorporating the resizing ratio and RGB feature space.
The Rank-1 accuracies are shown in ~\Cref{tab:CASIA-B_Rank}.
Compared to the original backbones, the models with prior knowledge consistently demonstrates an increase of 3.3\% for GaitSet on CL.
It is crucial to highlight that all improvements with GAR are accomplished without additional parameters involved but with an extra 1D convolution to extract ratio information in the test phase. 
\begin{table}[t]
\small
\setlength{\tabcolsep}{5pt}
\centering
\begin{tabular}{  c |c|c |c| l }
\toprule
Models & NM & BG & CL & Mean \\
\midrule
GaitSet~\cite{chao2019gaitset}& 89.7 & 77.6 &  59.3 &  75.5\\
 \rowcolor{GAR} $\hookrightarrow$w / GAR & 90.8 & 81.5 & 62.4& 78.2 \small \textcolor{green}{$\uparrow$ 2.7}\\
\midrule
GaitPart~\cite{fan2020gaitpart}& 90.5 & 78.7 &  64.2 &  77.8\\
 \rowcolor{GAR} $\hookrightarrow$w / GAR & 91.5 & 83.0 & 69.3& 81.2 \small \textcolor{green}{$\uparrow$ 3.4}\\
 \midrule

GaitBase~\cite{fan2023opengait} & 91.1 & 82.8 & 65.3 & 79.7 \\
\rowcolor{GAR} $\hookrightarrow$w / GAR & \textbf{92.3} & \textbf{85.2} & 67.7  & 81.7 \small \textcolor{green}{$\uparrow$ 2.0}\\
\midrule

 GaitGL~\cite{lin2021gait}& 91.1 & 82.7 & 66.6 & 80.1\\

\rowcolor{GAR} $\hookrightarrow$w / GAR & 91.5 & 83.1 & 67.9 & 80.8 \small \textcolor{green}{$\uparrow$ 0.7}\\

\bottomrule
\end{tabular}
\caption{Verification (TPR(\%)@FPR=$1e^{-2}$) on  CASIA-B.}
\vspace{-2mm}
\label{tab:CASIA_veri}
\end{table}

We also evaluate the \textit{verification} performance. The TAR(\%) results are shown in ~\Cref{tab:CASIA_veri}. Across the four backbones, GAR contributes to enhancements of 2.7\%, 3.4\%, 2.0\% and 0.7\%, attaining verification rates of 78.2\%, 81.2\%, 81.7\% and 80.8\%, respectively. 
It is noteworthy that although the recognition results in NM and BG approach saturation, there remains room for improvement in the verification task.

\noindent\textbf{Evaluation on Gait3D}~\cite{zheng2022gait} To evaluate our model on a public outdoor dataset, we also did cross-domain evaluation on Gait3D. The results are in ~\Cref{tab:gait3d}. We see that our proposed method achieves higher performance in all criteria. Especially, Rank-1 increases 2.0\% to 23.5\%. Since cross-domain evaluation is a challenging task, the results are lower than single-domain ones. 
The model trained on the BRIAR dataset exhibits superior performance compared to others, indicating that the model learned from the BRIAR dataset generalize well.

\noindent\textbf{Gait Detector Evaluation} We trained a gait detector using the DHSs generated from the BRIAR training set, and it reaches \underline{91.9\%}, \underline{86.1\%} and \underline{88.5\%} accuracy on BRIAR, CASIA-B and Gait3D respectively. For BRIAR, the false positive, true positive, false
negative and true negative for the gait detection module are 4.5\%, 53.1\%, 3.4\% and
38.8\%, respectively. The high accuracies show that the gait detector is robust to different domains. If the whole DHS is fed to the detector to get the prediction, the detection accuracy would be 82.7\%, 85.3\% and 87.1\%, respectively. The drop demonstrates the necessity of taking the split signature as an input. These results are under the assumption that the clips in curated datasets do not contain segments without gait information or incomplete body.

\subsection{Ablation Study}
\label{sec:ablation}

To show the impact of each part in our design, we conduct a series of ablation experiments.

\noindent\textbf{Ratio and RGB help better silhouette embedding}. In the gait recognition model, we evaluate the \emph{ratio attention} and \emph{cross modality distillation} by employing GaitGL as the backbone. 
From \Cref{tab:CASIA_ablation}, when we apply cross modality distillation, the recognition accuracy on CL reaches 85.0\% and its verification result increases by 0.5\% using the exact same model as GaitGL. As for ratio attention, it improves verification and Rank-1 accuracy under all conditions. Compared to the baseline, our proposed GAR gains a remarkable improvement on verification and Rank-1 in CL from 80.1\% and 83.3\% to 80.8\% and 85.7\% respectively. These improvements are also shown in the BRIAR dataset, which means the view angle cue from ratio and RGB's feature space help build a representative embedding.

\begin{table}[t]
\setlength{\tabcolsep}{1.5pt}

\centering
\small
\begin{tabular}{  c| c |c |c |c  |c | c }
\toprule
\footnotesize Source &\footnotesize Methods & \footnotesize Rank-1 & \footnotesize Rank-5 & \footnotesize Rank-10 & \footnotesize mAP & \footnotesize mINP  \\
\midrule
\multirow{2}{*}{CASIA-B} 
 &GaitSet~\cite{chao2019gaitset} & 6.9 & 14.6  & - & 4.5 & - \\
&GaitGL~\cite{lin2021gait} & 8.8 & 15.7  & 18.8 & 5.5 & 3.1\\
\midrule
\multirow{2}{*}{OU-MVLP} 
 &GaitSet~\cite{chao2019gaitset} & 6.1 & 12.4  &- & 4.4 & -\\
 &GaitGL~\cite{lin2021gait} & 16.4 & 25.8  & 31.2 & 13.1 & 7.3\\
\midrule
\multirow{2}{*}{GREW} 
 & GaitSet~\cite{chao2019gaitset} & 16.5 & 31.1  & - & 11.7 & - \\
 &GaitGL~\cite{lin2021gait} & 18.3 & 31.9  & 39.2 & 13.1 & 7.3 \\
\midrule

 &GaitGL~\cite{lin2021gait} &  21.5 & 36.5  & 42.5 & 15.2 & 8.2\\
\rowcolor{GAR} \cellcolor[rgb]{1,1,1}BRIAR & $\hookrightarrow$w / GAR & 23.5  & 37.4  & 43.4 & 15.9 & 8.2 \\

\bottomrule

\end{tabular}%

\caption{Cross domain evaluation on Gait3D with detector.}
\label{tab:gait3d}
\vspace{-2mm}

\end{table}

\noindent\textbf{RGB modality is sensitive to appearance change}. In \Cref{tab:reid}, we evaluate the framework only using the RGB modality with GaitGL as a feature extraction model, i.e. GaitGL$_{RGB}$. We observe that it has lower performance than silhouette-based one, i.e. GaitGL, decreasing from 17.3\% to 15.6\%. Considering GaitGL is a model focusing on temporal rather than spatial information, we further experiment on CAL~\cite{Gu_2022_CVPR}, which achieves SoTA in public clothes-changing ReID datasets. But it only reaches 17.6\%, still lower than gait-based methods. So, as mentioned in~\cite{shen2022comprehensive}, even though RGB is an acceptable modality for gait recognition, it is not widely applied due to its sensitivity to appearance change. However, its feature has unique gait information that augments the silhouette feature.

\noindent\textbf{Mixing gait and non-walking adversely affect feature aggregation}.
To demonstrate the effectiveness of the gait detector, we first train with all BRIAR training set data, i.e. GaitGL w/ stand, including standing, random walking, and structure walking. From \Cref{tab:reid}, we see that when standing sequences are included in the training process, the gait recognition performance drops, which means that static sequences corrupt the gait embedding construction.
In \Cref{tab:BRIAR_rank_veri}, we show the improved performance obtained by applying GADER--GAR on segments that contain gait information with a complete body while employing SemReID~\cite{huang2023selfsupervised} on remaining data. GADER's superior performance shows that it is able to process segments with gait information with a complete body rather well.
And this improvement is achieved with little cost, a lightweight classification network, since the DHS is extracted from the input provided to gait recognition. What is more, the well-trained model is robust to domain gaps among different datasets, so it can be directly applied.

\begin{table}[t]
\small
\begin{subtable}[t]{.55\linewidth}
\centering
\setlength{\tabcolsep}{0.2pt}
    \begin{tabular}{  c c |c c c c}
    
    \hline
    Ratio & Cross & $R_{NM}$ & $R_{BG}$& $R_{CL}$& $Veri$\\
    
    \hline
    \xmark & \xmark&  97.2 & 94.1 & 83.3 &80.1\\
    \cmark& \xmark&  97.4 & 94.4 &85.1 & 80.4\\
    \xmark& \cmark&  97.3 & \textbf{94.5} & 85.0 & 80.6\\
    
    \cmark & \cmark & \textbf{97.5} & \textbf{94.5} & \textbf{85.7} & \textbf{80.8}\\

    \hline
    \end{tabular}
    \caption{CASIA-B: recognition accuracy of three probes ($R_{NM}$, $R_{BG}$, $R_{CL}$) and average verification ($Veri$).}
\end{subtable}
\hspace{0.1in}
\begin{subtable}[t]{.4\linewidth}
\raggedright
\setlength{\tabcolsep}{0.2pt}
    \begin{tabular}{  c c |c c}
    \hline
    Ratio & Cross &  Rank-1& $Veri$\\
    
    \hline
    \xmark & \xmark&  17.3 & 33.1\\
    \cmark& \xmark&  26.3 & 33.4\\
    \xmark& \cmark&  25.2 & 33.3\\
    
    \cmark & \cmark & \textbf{28.3} & \textbf{33.6}\\

    \hline
    \end{tabular}
    \caption{BRIAR: mean rank-1 accuracy(Rank-1) and verification ($Veri$).}
\end{subtable}

\caption{Ablation studies on ratio attention and cross-modality distillation. The results are shown using recognition accuracy and verification (TPR(\%)@FPR=$1e^{-2}$) on CASIA-B and BRIAR.}
\label{tab:CASIA_ablation}
\vspace{-3mm}
\end{table}

\begin{table}[t]

\centering
\small
\begin{tabular}{ c| c | c |c}
    \hline
    \small Modality &\small Models & \small Rank-1 & \small Rank-1 w/detect\\
    \hline

\multirow{2}{*}{RGB} 
&\small {CAL}\cite{Gu_2022_CVPR} &  \small 17.6 & \small 18.3\\
    &\small GaitGL$_{RGB}$ &  \small 15.6 & \small 14.8\\
    
    \hline

\multirow{3}{*}{Silhouette} &
    \small GaitGL w/stand &  \small 24.2 & \small 26.4 \\
    &\small GaitGL\cite{lin2021gait} &  \small 17.3 & \small 29.8\\

\rowcolor{GAR} \cellcolor[rgb]{1,1,1}    &$\hookrightarrow$w / GAR & \small 28.3 & \small \textbf{42.5}\\
    \hline
    \end{tabular}
    \caption{Comparison among different models tested on full set and detected qualified gait set through gait detection on BRIAR.}
\vspace{-5mm}
\label{tab:reid}
\end{table}

\subsection{Failure Cases}
\label{sec:detectfail}

\begin{figure}[htb]
\vspace{-.3cm}
    \setlength{\abovecaptionskip}{3pt}
    \setlength{\tabcolsep}{2pt}
    \centering
    \begin{tabular}[b]{ccc}
        \begin{subfigure}[b]{.3\linewidth}
            \includegraphics[width= \textwidth,height=.7\textwidth]{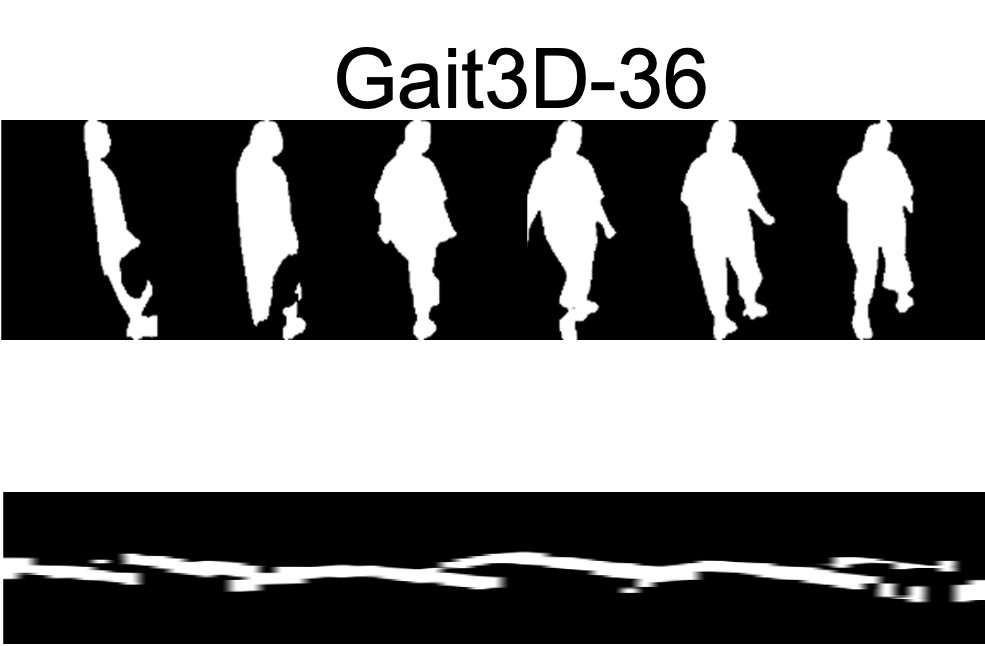}

        \end{subfigure} &
        \begin{subfigure}[b]{.3\linewidth}
            \includegraphics[width= \textwidth,height=.7\textwidth]{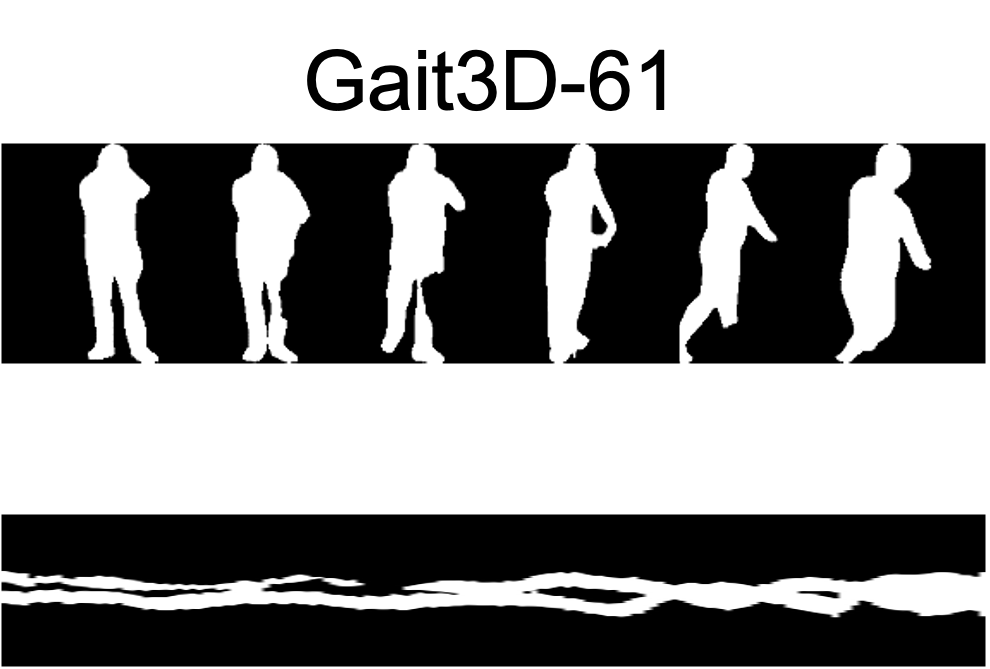}
        \end{subfigure} &
        \begin{subfigure}[b]{.3\linewidth}
            \includegraphics[width= \textwidth,height=.7\textwidth]{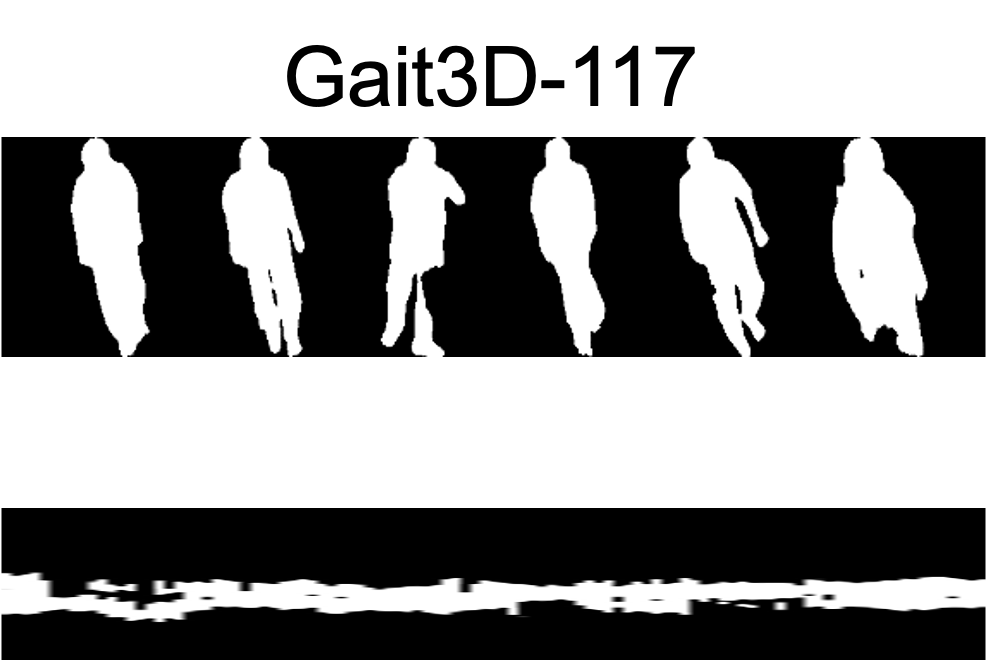}

        \end{subfigure} \\
        \begin{subfigure}[b]{.3\linewidth}
            \includegraphics[width= \textwidth,height=.7\textwidth]{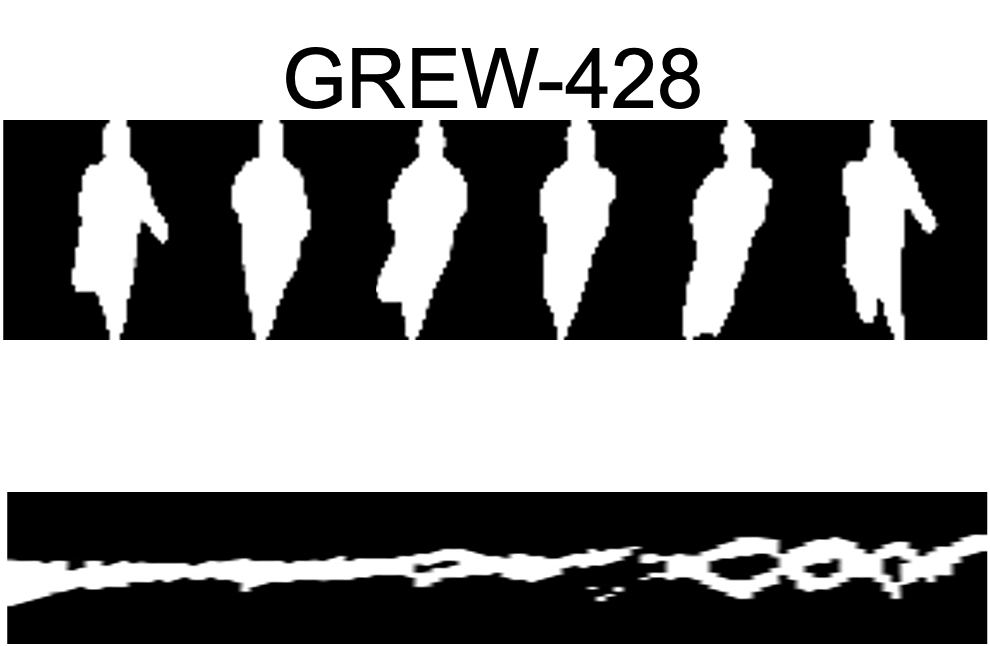}

        \end{subfigure} &
        \begin{subfigure}[b]{.3\linewidth}
            \includegraphics[width= \textwidth,height=.7\textwidth]{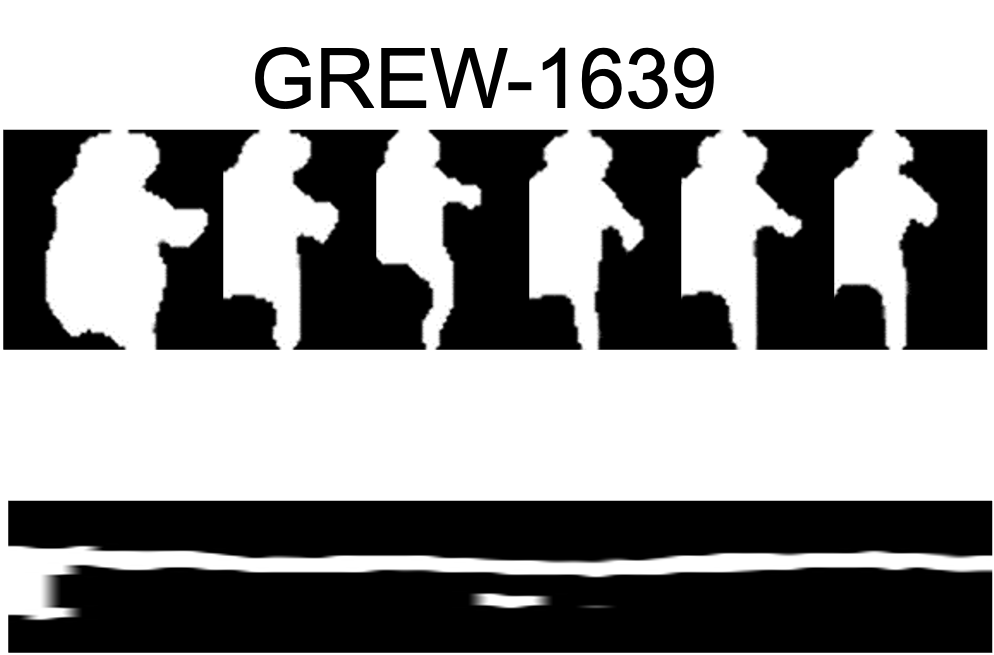}
        \end{subfigure} &
        \begin{subfigure}[b]{.3\linewidth}
            \includegraphics[width= \textwidth,height=.7\textwidth]{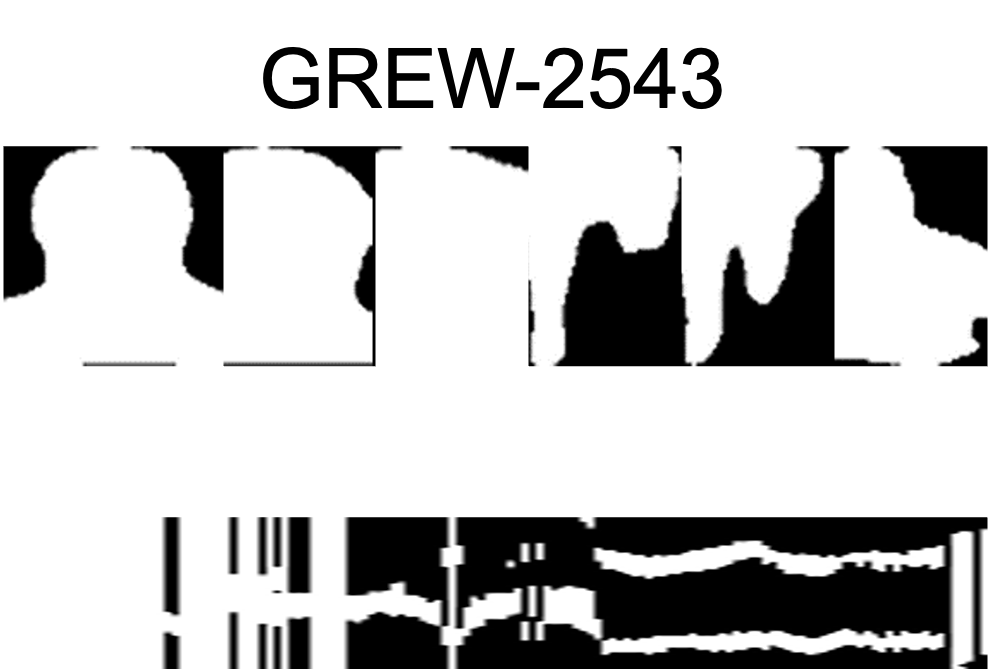}

        \end{subfigure}

    \end{tabular}
    \caption{Some examples showing the absence of human movement in videos appearing in the curated dataset. We record the dataset name and corresponding subject id of the sequence.}
   \label{fig:curated}
   \vspace{-2mm}
\end{figure}

\noindent\textbf{Segments without gait appear even in curated datasets.}
We recognize the presence of incomplete body segments or sequences without discernible movement in the BRIAR dataset, simulating real-life gait recognition challenges. 
But with closer examination, we found this issue also exists in curated datsets. We show some examples from GREW~\cite{zhu2021gait} and Gait3D~\cite{zheng2022gait} in \cref{fig:curated}. 
These examples emphasize the need for the gait detector even for curated datasets

\begin{figure}[htb]
    \setlength{\abovecaptionskip}{3pt}
    \setlength{\tabcolsep}{1pt}
    \centering
    \begin{tabular}[b]{cc}
    \captionsetup[subfigure]{justification=centering}
        \begin{subfigure}[b]{.45\linewidth}
            \includegraphics[width= \textwidth,height=0.2\textwidth]{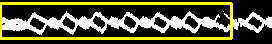}
            \caption*{\footnotesize GT : full-gait  \xspace \xspace \xspace \xspace
            P : \textcolor{green}{full-gait} }

        \end{subfigure} &
    \captionsetup[subfigure]{justification=centering}
        \begin{subfigure}[b]{.45\linewidth}
            \includegraphics[width= \textwidth,height=0.2\textwidth]{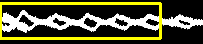}
            \caption*{\footnotesize GT : full-gait  \xspace \xspace \xspace \xspace
            P : \textcolor{green}{full-gait} }
        \end{subfigure}\\
    \captionsetup[subfigure]{justification=centering}
        \begin{subfigure}[b]{.45\linewidth}
            \includegraphics[width= \textwidth,height=0.2\textwidth]{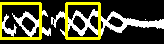}
            \caption*{\footnotesize GT : full-gait  \xspace \xspace \xspace \xspace
            P : \textcolor{red}{full-stand} }

        \end{subfigure} &
    \captionsetup[subfigure]{justification=centering}
        \begin{subfigure}[b]{.45\linewidth}
            \includegraphics[width= \textwidth,height=0.2\textwidth]{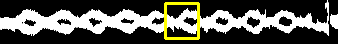}
            \caption*{\footnotesize GT : full-gait  \xspace \xspace \xspace \xspace
            P: \textcolor{red}{part-gait} }
        \end{subfigure}

    \end{tabular}
    \caption{Some gait detector failure cases.}
   \label{fig:fail}
   \vspace{-.5cm}
\end{figure}

\noindent\textbf{Analysis}
During test time, a fixed window length is used, which can result in \textit{missing} gait sequences or non-gait segments involved, as demonstrated in~\cref{fig:fail} first row, but the majority of the segments containing gait information are still caught. Furthermore, since we only have annotations for the entire DHS sequences rather than individual moments, it is challenging to train a robust model to always make correct predictions especially when the appearance is misleading, as shown in~\cref{fig:fail} second row. Despite these challenges, our gait detector can accurately identify frames that contain gait signatures.

\vspace{-.1cm}

\vspace{-0.2cm}
\section{Conclusion}
\label{sec:Conclusion}
\vspace{-0.1cm}
In this paper, we present a novel gait detection and recognition approach to address challenging unconstrained conditions. First, we introduce a gait detector to identify frames that contain gait with a complete body. 
With the help of a gait detector, gait recognition and person ReID can cooperate complementarily to achieve higher performance. Secondly, the gait recognition pipeline utilizes both RGB and silhouette modality to learn robust representations. Notably, we fill the viewpoint information leakage with a simple yet effective ratio attention signal.
Additionally, we enhance the silhouette modality embedding through feature distillation from the RGB modality. Such a design helps to leverage the well-learned feature space of RGB modality with the robustness of silhouettes and does not require RGB data at test time. Through extensive experiments, we show that our proposed method improves the performance on Gait3D in cross-domain evaluation and achieves SoTA performance in the standard CASIA-B and the challenging BRIAR dataset. %

\section{Acknowledgement}

This research is based upon work supported in part by
the Office of the Director of National Intelligence (ODNI),
Intelligence Advanced Research Projects Activity (IARPA),
via [2022-21102100005]. The views and conclusions contained
herein are those of the authors and should not be
interpreted as necessarily representing the official policies,
either expressed or implied, of ODNI, IARPA, or the U.S.
Government. The US. Government is authorized to reproduce
and distribute reprints for governmental purposes notwithstanding
any copyright annotation therein.

\newpage

{\small
\bibliographystyle{ieee}
\bibliography{egbib}
}

\end{document}